\documentclass[10pt,twocolumn,letterpaper]{article}

\usepackage{iccv}
\usepackage{times}
\usepackage{epsfig}
\usepackage{graphicx}
\usepackage{amsmath}
\usepackage{amssymb}
\usepackage{amsthm}
\usepackage{algpseudocode}
\usepackage{algorithm}
\usepackage{array}
\usepackage{caption}
\usepackage{subcaption}

\usepackage[pagebackref=true,breaklinks=true,letterpaper=true,colorlinks,bookmarks=false]{hyperref}

\iccvfinalcopy 

\def\iccvPaperID{} 

\ificcvfinal\fi
\begin{document}

\title{First-Take-All: Temporal Order-Preserving Hashing for 3D Action Videos}

\author{Jun Ye\\
University of Central Florida\\
{\tt\small jye@cs.ucf.edu}
\and
Hao Hu\\
University of Central Florida\\
{\tt\small hao\_hu@knights.ucf.edu}
\and
Kai Li\\
University of Central Florida\\
{\tt\small kaili@cs.ucf.edu}
\and
Guo-Jun Qi \thanks{Guo-Jun Qi is the corresponding author}\\
University of Central Florida\\
{\tt\small guojun.qi@ucf.edu}
\and
Kien A. Hua\\
University of Central Florida\\
{\tt\small kienhua@eecs.ucf.edu}
}

\maketitle

\begin{abstract}
With the prevalence of the commodity depth cameras, the new paradigm of user interfaces based on 3D motion capturing and recognition have dramatically changed the way of interactions between human and computers. Human action recognition, as one of the key components in these devices, plays an important role to guarantee the quality of user experience. Although the model-driven methods have achieved huge success, they cannot provide a scalable solution for efficiently storing, retrieving and recognizing actions in the large-scale applications. These models are also vulnerable to the temporal translation and warping, as well as the variations in motion scales and execution rates. To address these challenges, we propose to treat the 3D human action recognition as a video-level hashing problem and propose a novel First-Take-All (FTA) Hashing algorithm capable of hashing the entire video into hash codes of fixed length. We demonstrate that this FTA algorithm produces a compact representation of the video invariant to the above mentioned variations, through which action recognition can be solved by an efficient nearest neighbor search by the Hamming distance between the FTA hash codes. Experiments on the public 3D human action datasets shows that the FTA algorithm can reach a recognition accuracy higher than 80\%, with about $15$ bits per frame considering there are $65$ frames per video over the datasets.
\end{abstract}

\section{Introduction}
The recent advances in the commodity depth sensors such as Microsoft Kinect, Intel RealSense and LeapMotion have dramatically changed the way of human-computer interaction. The new generation of user interfaces based on 3D motion capturing and recognition make the interactions between humans and computers easier than ever before. These interfaces have already enabled a wide range of applications including video games, education, business and healthcare. Behind all these applications, the 3D human action recognition plays a key role and directly determines the quality of the user experience.

Although a great number of works \cite{muller2006motion, wang2012mining, hon4d, 2014snv, 2014Liegroup, xia2012view} have been developed for solving the problem of automatic human action recognition,  the modeling of dynamic structures of human actions remains challenging due to the temporal translation and warping of the action sequences, as well as the variation in the motion scales and the execution rates of the actions \cite{zhao2013online}. More importantly, the current model-driven solutions normally require a dedicated classifier for each class of actions and cannot provide a scalable solution for the large-scale action recognition applications.

Inspired by the success of the hashing techniques in image retrieval \cite{LSH1}, we treat the 3D human action recognition as a hashing problem of encoding videos with compact binary sequences of fixed length, so that the similarity between videos can be compared by the Hamming distance between their hash codes preserving the intrinsic temporal structure of actions. Thus, action recognition can be solved by an efficient approximate nearest neighbor search based on the hash codes of videos. Most of the existing hashing algorithms \cite{LSH1, LSH3, minHas, compressedHash} are developed for images with the fixed resolution/dimension.

We note there exist some hashing algorithm \cite{ye2013large} that handles the large-scale video datasets while considering the temporal consistency. However, such method still applies the hashing to each individual frame rather than the entire sequence.  Hence, to measure the video similarity, we have to compute the average similarity between each pair of frames, which can be computationally prohibitive especially with a ever-growing length of videos in many applications. To address this problem, we propose to hash the entire video as a whole into the bit sequence of fixed length to facilitate direct computation of Hamming distance.  There are two challenges facing the hashing of the entire video: (1) encoding the temporal structures of actions and (2) dealing with the varying length of videos to generate fixed length of hash codes.  Ideally, we wish that the generated hash codes can be resilient against temporal translation and warping, variations in scales and execution rates.

To address the above challenges, we propose a novel temporal order-preserving hashing algorithm, namely First-Take-All (FTA). The FTA hashing algorithm first applies multiple random projections to translate a video into several sequences of latent postures. Then it encodes the video by the temporal order of the occurrence of these postures.  Specifically, in each iteration that generates a new hash code, a group of $k$ latent postures are randomly selected, and then a video is encoded by the index of the posture that is acted first.  After several iterations, a set of hash codes generated in this FTA fashion can capture the temporal order of latent postures acted in a video, and the similarity between videos can be measured by computing the Hamming distance between the FTA hash codes. Since the hashing is applied to the entire video, it can normally achieve a low bit rate per frame, making it much efficient compared with the model-based approaches.  In addition, we will show that the FTA hashing is invariant to the temporal translation and warping, as well as the variations in motion scales and execution rates, as long as the temporal-order of the posture sequence does not change for a class of actions.  This makes FTA robust against the intra and inter-class variations caused by individual actors.

The FTA hashing algorithm extends the Winner-Take-All (WTA) algorithm \cite{yagnik2011power}, but differs from it in several significant aspects.  First of all, WTA is not an algorithm that can be applied to hashing varied length of sequences.  In fact, it must assume that all the input vectors reside in a feature space of fixed dimension. For this reason, WTA has only been applied to hash the data of fixed length like images and text. Second, WTA compares the order of features chosen from the original space.  This unnecessarily limits its ability in capturing the ranking structure among various subspaces.  On the contrary, the proposed FTA hashing is more expressive in representing the temporal order of latent postures obtained by projecting the sequence into several subspaces.


The main contributions of the paper are:
\begin{enumerate}
\item We propose a novel FTA algorithm for hashing videos of varied length. The hashing algorithm is invariant to temporal translation, scale variation and execution rate variation; and
\item We perform extensive experiment studies on three public 3D human action datasets and demonstrate the performance of the proposed FTA Hashing algorithm by comparing it with a baseline method without leveraging the temporal-order information.
\end{enumerate}

To the best of our knowledge, this is the first work to perform hashing on the entire video sequence of varied length for the recognition of human actions. It is also worth noting that, the proposed FTA hashing is not limited to human action videos, it can also be potentially applied as a generic temporal hashing algorithm to other types of video sequences. 

The remainder of the paper is organized as follows.  Section 2 briefly reviews the related work.  The First-Take-All hashing algorithm is introduced and discussed in Section 3. Experiments and performance study are presented in Section 4. Finally, Section 5 concludes the paper.


\section{Related work}
We review the related works in two following categories.
\subsection*{Human Action Recognition and Retrieval}
Modeling temporal structure of video sequences is one of the most challenging problems in human action recognition and has attracted intensive research. Many of the existing approaches focus on extracting the local spatio-temporal features and do not explicitly model the temporal patterns of the action sequence. Most of these works are histogram-based and adopting the bag-of-words framework. For example, in \cite{li2010action}, the bag-of-3D-points from the depth maps are sampled and clustered to model the dynamics of human actions. Similar ideas are presented in \cite{yang2012recognizing}, where the Histogram of Gradient (HoG) features are exacted from the depth motion maps to classify human actions.  Histogram of 3D joints (HOJ3D) \cite{xia2012view} and histogram of visual words \cite{2013cvprw1} are also employed to describe the action sequences by using the joint features. Unfortunately, these histogram-based methods do not preserve the temporal order of the primitive postures of the action and may lead to the poor performance on distinguishing between actions composed of the similar postures but in different temporal orders.

It is obvious that the temporal characteristics of human actions must be fully explored in order to achieve a high recognition rate. Motion template-based approaches \cite{muller2006motion, zhao2013online} are introduced to model the temporal dynamics of actions, where the Dynamic Time Warping (DTW) algorithm is used to align the sequences of varied length and execution rate. On the contrary, the Temporal Pyramid \cite{wang2012mining, hon4d} attempts to represent the temporal structure of the sequence by uniformly subdividing the sequence into several partitions. With the uniform temporal partition, however, the temporal pyramid is less flexible to handle the execution rate variation. The Adaptive Temporal Pyramid \cite{2014snv} is instead proposed to overcome this problem by adaptively dividing the temporal sequence by the motion energy.  Vemulapalli et al. \cite{2014Liegroup} presents a body-part representation of the human skeleton, where the temporal dynamics in terms of 3D transformation are projected onto a curved manifold in the Lie group.



\subsection*{Hashing Algorithms}
Hashing algorithms are widely adopted in the approximate nearest neighbor search problem \cite{LSH1,LSH3}. There are plenty of works aiming at achieving a higher retrieval rate with shorter code length.  We categorize these existing hashing algorithms into two families -- the space-partitioning methods and the ranking-based methods.

The space-partitioning methods, such as Locality Sensitive Hashing (LSH) \cite{LSH1} and Compressed Hashing \cite{compressedHash} normally partition the whole feature space into a sequence of half spaces and quantize the original features into binary bits by these half spaces. In order to preserve the Euclidean distances between high dimensional vectors with sufficient precision, long codes are usually required for these methods. To overcome this drawback, Multi-Probe LSH \cite{Lv2007} and entropy-based LSH \cite{Panigrahy2006} are proposed to reduce the storage burden at the cost of increasing the query time. On the other hand, different paradigms of hashing algorithms have been developed to approximate distance metrics other than the Euclidean distance, such as the $p$-norm distance\cite{LSH3} and the Mahalanobis distance \cite{kulis2009fast}. Spherical LSH \cite{terasawa2007spherical} works for hashing a set of points on the hypersphere of an input space. There also exist several works kernelizing the LSH approaches by considering the Reproducing Kernel Hilbert Space (RKHS) \cite{kulis2009kernelized, raginsky2009locality}.

Unlike space-partitioning methods, the ranking-based hashing methods encode the ordinal relation between the original features rather than their magnitudes.  For example, Min Hashing \cite{minHas} approximates the Jaccard similarity coefficient between two sets by encoding a set with the minimal value of a hash function over its members. Recently, the Winner-Take-All (WTA) Hash \cite{yagnik2011power} has been proposed to encode the magnitude orders of randomly permutated features. The resultant hash codes are scale-invariant, and are often more resilient against the noises. Some other ranking-based hashing algorithms, like the Rank-Sensitive Hash \cite{RSH}, can be regarded as a special case of the WTA when the window size of ordinal comparison is $2$.

However, to the best of our knowledge, there exists no hashing algorithm aiming at encoding the temporal structure of entire sequences of varied length.  Ye et al. \cite{ye2013large} introduce a supervised hashing algorithm for video sequences. However, it still performs hashing on individual frames, rather than on the entire video. Then the video similarity has to be computed by the average Hamming distance between each pair of frames. In contrast, we attempt to expand the scope of ranking-based hashing methods to directly explore the temporal order of actions on the video level.  This can yield much compact hash codes for videos, and the similarity between videos can be directly computed by the Hamming distance between the video-level hash codes.

\section{Temporal Order-Preserving Hashing}

In order to hash human action sequences into binary bits while preserving their temporal order, we
apply random projection to a video. This generates a sequence of confidence scores measuring whether each frame of the video belonging to a (unlabeled) posture corresponding to the random projection.
We generate several random projections that are applied to a video, and encode the video with the index of the random projection of which the peak of the confidence score comes the first.
This is why our algorithm is name First-Take-All (FTA).

This process will be repeated several times, and a sequence of hashing codes will be generated for a video.
In this fashion, the temporal order in which the postures are performed in a video will be encoded.
In Section \ref{fd}, we formalize the algorithm, and then we will explain the intuitive idea behind the formal description.
\subsection{Formal Description} \label{fd}

In this section, we formally present the proposed temporal order-preserving hashing algorithm for video sequence -- First-Take-All (FTA).

\subsubsection*{Random Projections for Latent Postures}
The method can be formulated as follows. Suppose a video of length $n$ is represented as a sequence of frames $\mathbf X = [\mathbf x_1, \mathbf x_2\, \cdots, \mathbf x_n]$.  Each frame $\mathbf x_i \in \mathbb R^d$, $i=1,\cdots,n$, is represented by a $d$-dimensional feature vector.  First, we generate $m$ random projections $\mathbf W=[\mathbf w_1, \mathbf w_2, \cdots, \mathbf w_m]$, each $\mathbf w_l\in\mathbb R^d$ is drawn from a multivariate Gaussian distribution, i.e., $\mathbf w_l \sim \mathcal N(\mathbf 0,\sigma^2\mathbf I)$ for $l=1,\cdots,m$.

As aforementioned, each random projection can be interpreted as forming a linear subspace for a unknown posture. Hence, the inner product $S_{l,i}=\mathbf w_l^\intercal \mathbf x_i$ represents the confidence score of posture $l$ on frame $i$. In a compact matrix form, we can use a matrix of size $m\times n$
$$
\mathbf S \triangleq [S_{l,i}]_{m\times n} = \mathbf W^\intercal X
$$
each row $l$ of which represents the application of random projection $\mathbf w_l$ to the entire video sequence.

\subsubsection*{First-Take-All}
Our goal is to design a hashing coding mechanism to preserve the temporal order structure. With the posture sequence in each row of the resultant matrix $\mathbf S$, we can find which posture is first performed to encode the video. Formally, we randomly choose $k$ rows indexed by $\mathcal K =\{l_1, \cdots, l_k\}$ from $\mathbf S$, each row representing the sequence of a latent posture.  When $k=2$, it is a pair-wise comparison of the temporal orders between two postures. Otherwise, when $k>2$, the comparison is made between multiple postures.

To decide which of postures comes first, we need to test when each posture is performed for the first time in the sequence.  Here we introduce two approaches.

{\noindent\bf FTA by peak} The first approach is to use the peak of confidence score to denote the first time a posture is acted.  Suppose for posture $l_j\in \mathcal K$, its peak is attained at frame $i_{l_j}$, i.e.,
$$
i_{l_j} = \arg\max\limits_{1\leq i\leq n}  \{S_{l_j,i}|S_{l_j,i}\geq \theta\}
$$
So $i_{l_j}$ denotes the time when the confidence that posture $l_j$ is performed reaches the peak.  Note that here we require that the confidence score $\mathbf S_{l_j,i}$ should be larger than a threshold $\theta$ for a frame $i$ to be considered as posture $l_j$.  This is used to rule out those less salient postures and provide a more robust result for the following temporal order comparison. If the confidence score for posture $l_j$ has never passed this threshold, it implies the posture might have never been performed.   In this case, we set $i_{l_j}$ to $+\infty$ by convention.

Then the posture that comes first is given by
\begin{equation}\label{code}
j^\star = \arg\min\limits_{1\leq j \leq k} i_{l_j},~{\rm s.t.}~l_j\in\mathcal K
\end{equation}
where $\mathcal K$ is the selected rows for the current iteration. We will use $j^\star$ as the hash code for the video, which denotes the index of the posture first performed in the sequence. If all $i_{l_j},l_j\in\mathcal K$ involved are $+\infty$ (i.e., none of postures have been performed), the above temporal-order comparison fails, and we output a special code $0$ to denote this case.

\begin{figure}[t]
\begin{center}
   \includegraphics[width=0.7\linewidth]{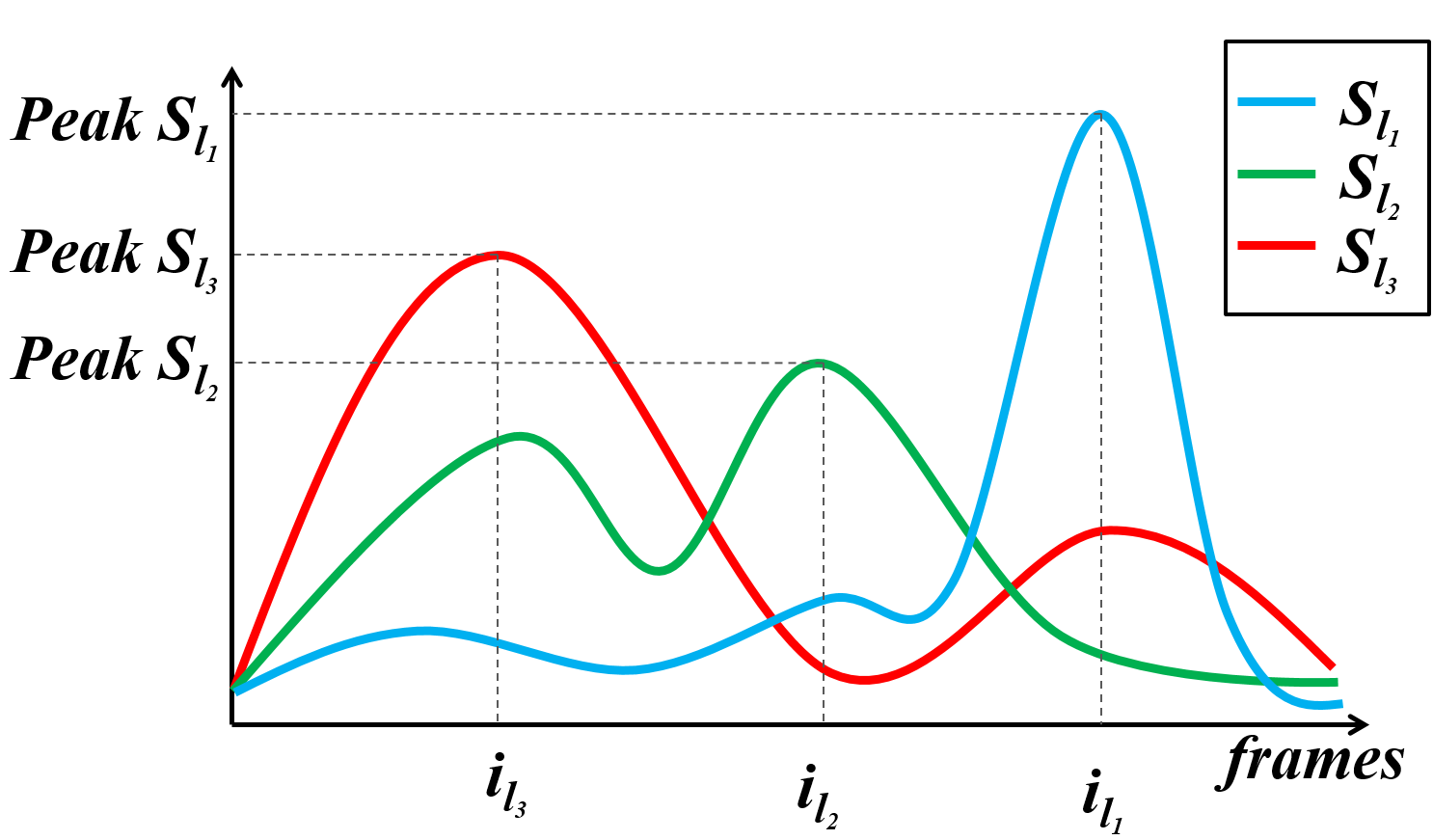}
\end{center}
   \caption{Illustration of the FTA by peak approach when $K = 3$}
\label{fig-explain}
\end{figure}

We illustrate an FTA by peak example in Figure~\ref{fig-explain}, where we plot three posture sequences ${\mathbf S_{l_1}, \mathbf S_{l_2}, \mathbf S_{l_3}}$, each corresponding to a row of matrix $\mathbf S$. For simplicity, we assume that the peak of three sequences passes the preset threshold $\theta$.
If we set $k=3$, all three sequences are involved, and the peak of $S_{l_3}$ comes at first and thus the code of the current iteration produces a hash code $3$. Otherwise, if we set $k=2$, a pair of sequences are randomly chose.  Suppose we choose $\mathcal K=\{l_1,l_2\}$. Because $i_{l_2}<i_{l_1}$, we output $j^\star=2$ to encode that the peak of posture sequence $l_2$ comes first.

FTA by peak is a much conservative approach to decide the temporal order between the postures.  Usually well before the peak, a posture has already been acted for a while.
In contrast to this conservative approach, we introduce a more aggressive approach below.

{\noindent\bf FTA by thresholding} In this approach, we assume that the first time a posture is performed in a sequence is when its confidence score reaches the preset threshold $\theta$ for the first time. Formally, we have
$$
i_{l_j} = \min \{i| S_{l_j,i}\geq\theta\}
$$
where $l_j\in\mathcal K$ belong to the selected postures.
By convention, we set $i_{l_j}$ to $+\infty$ if the above set is empty, denoting this posture $l$ has never been acted in the sequence.
Accordingly, the index of the posture first performed is also determined by Eq.~(\ref{code}).

In an extreme case that there is no posture from $\mathcal K$ passing the threshold test, all $i_{l_j}$ would be $+\infty$ for $l_j\in\mathcal K$.
 Then we produce a special hashing code $0$ to encode the video. Therefore,
in each iteration, we have a code book $\{0,1,\cdots,k\}$ with $k+1$ entries for the encoding of a video.
Apparently, a $(k+1)$-ary code can be represented by $\lceil\log (k+1)\rceil$ binary bits. We repeat the above coding iterations $p$ times, and we will get $p$ $(k+1)$-ary codes or equivalently $p\lceil\log (k+1)\rceil$ binary codes to encode an input sequence.

An algorithmic overview of our method is shown in Algorithm~\ref{TOPH}.

\begin{algorithm}
\caption{FTA Hashing}
\label{TOPH}
\begin{algorithmic}[1]

\Procedure{FTA}{$m$, $k$, $p$, $\mathbf W$, $X$, $\theta$}
    \State Generate $\mathbf W$ from Gaussian distribution;
    \State Set $\mathbf S \gets \mathbf W^\intercal \mathbf X$; 
    \State Initialize $\mathbf b$ as an empty binary sequence.
    \For{$\tau=1$ to $p$}  \label{iterBegin}
        \State Randomly select $k$ rows $\mathcal K$ from $\mathbf S$;
        \For{$j=1$ to $k$}
            \State Compute the first-acting time $i_{l_j}$ for $l_j\in\mathcal K$;
        \EndFor
        \State Set $j^\star \gets \arg\min\limits_{1\leq j\leq k} i_{l_j},~{\rm s.t.}~l_j\in\mathcal K$
        \State $\mathbf b \gets \mathbf b \cup \{j^\star\}$
	\EndFor \label{iterEnd}\\
	\Return $\mathbf b$
\EndProcedure
\end{algorithmic}
\end{algorithm}

Apparently, the complexity to compute one hash code is $O(k\log n+\log k)$, the total complexity for $p$ hash codes as well as the cost for the random projection is $O(p(k\log n+\log k)+mdn)$. Considering $k<<n$, the total complexity can be estimated as $O(p(k\log n)+mdn)$, which is linear with respect to the input arguments.

\subsection{The Invariance Properties of FTA}
\begin{figure*}
\begin{center}
\begin{subfigure}{0.33\textwidth}
\includegraphics[width=1.15\linewidth]{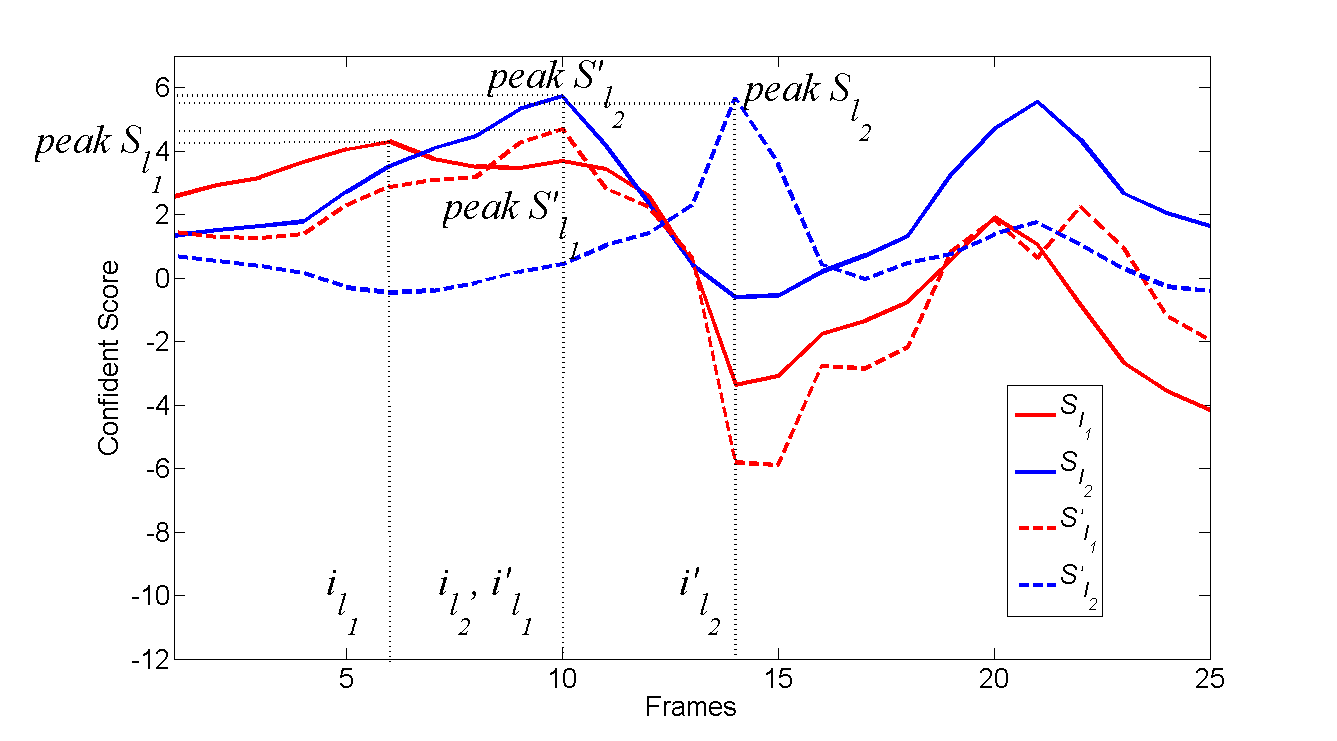}
        \caption{translation invariance}
        \label{fig:}
 \end{subfigure}
~
\begin{subfigure}{0.32\textwidth}
\includegraphics[width=1.15\linewidth]{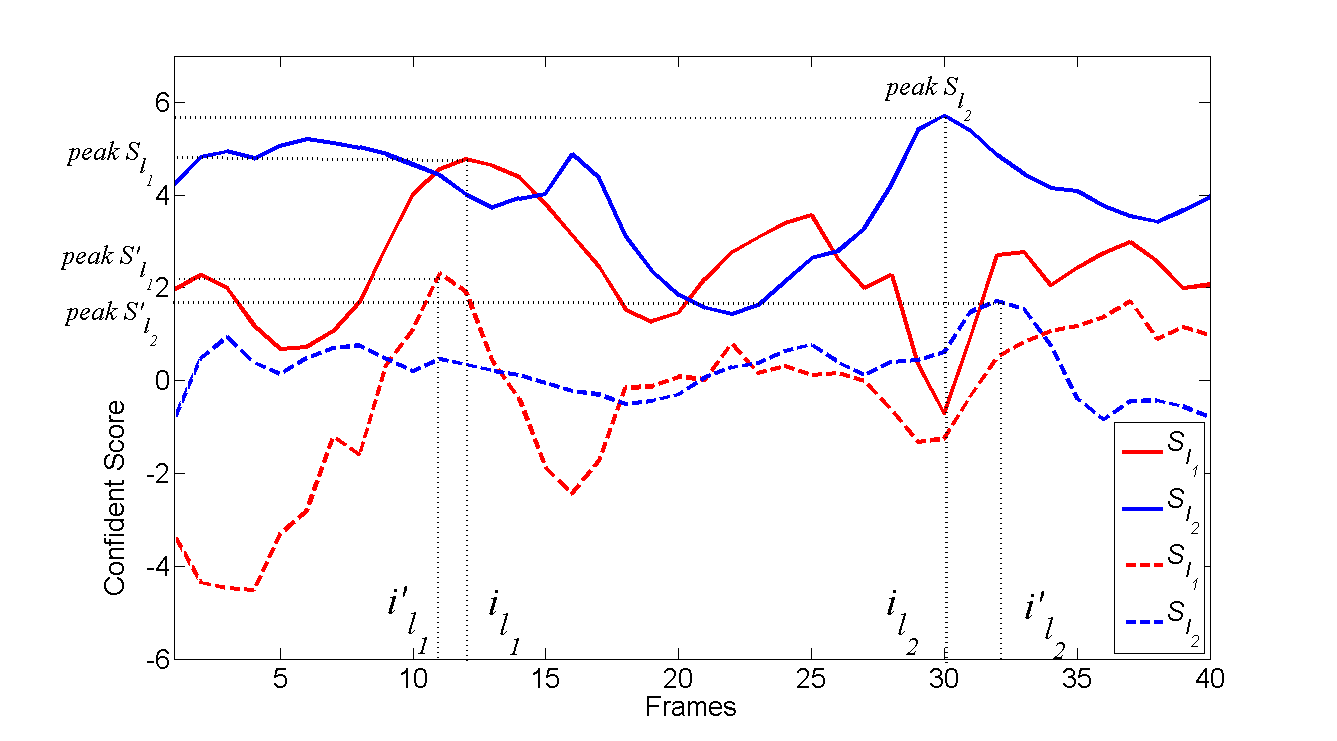}
        \caption{scale invariance}
        \label{fig:}
 \end{subfigure}
~
\begin{subfigure}{0.32\textwidth}
\includegraphics[width=1.15\linewidth]{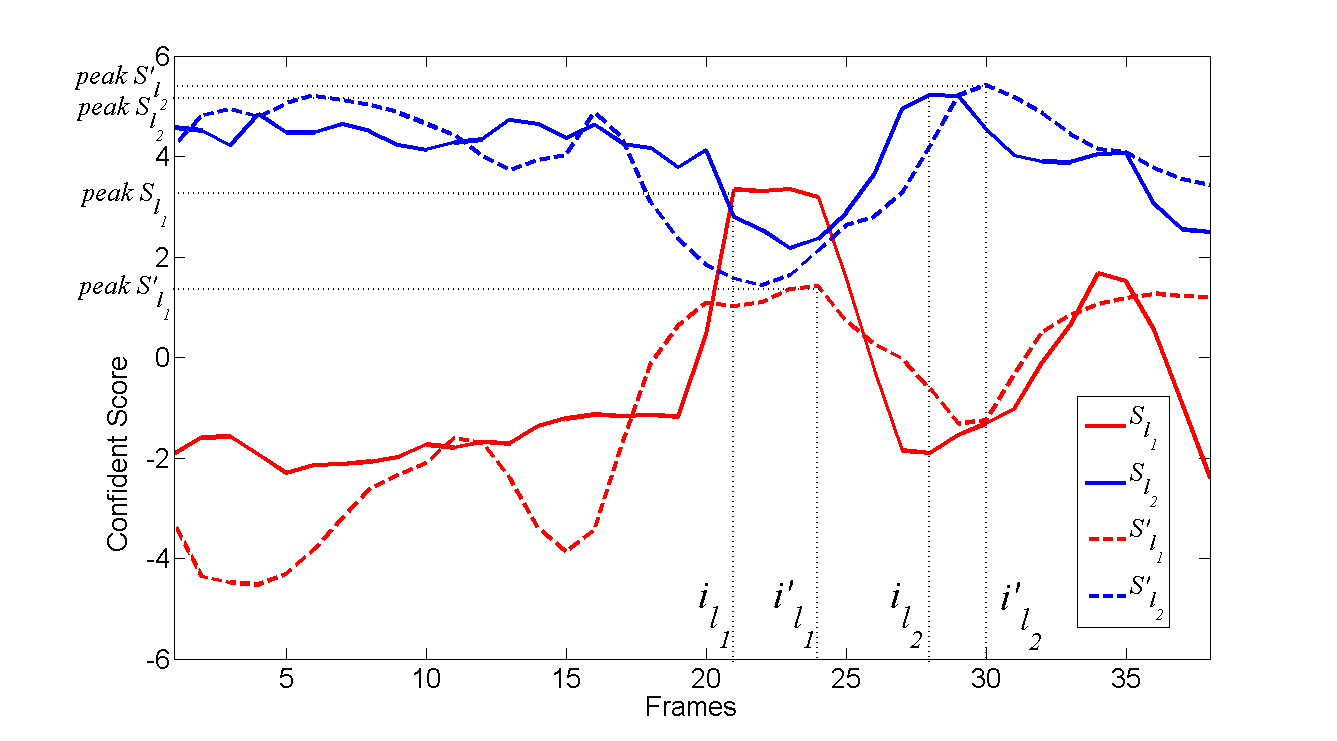}
        \caption{execution rate invariance}
        \label{fig:}
 \end{subfigure}
\end{center}
   \caption{Running examples to demonstrate the invariance properties of the FTA hashing. Two postures $l_1, l_2$ (red and blue) are investigated on two videos $\mathbf X, \mathbf X'$ (solid and dash) of the same action class. }
\label{fig-Invariance}
\end{figure*}

As mentioned in section 1, it is very challenging to model the temporal characteristics of action videos due to the temporal translation and warping as well as the variation in motion scales and execution rates. In this subsection, we show the nice properties of the FTA hashing, demonstrating that it is insensitive to the above variations by its temporal order-preserving nature.

Figure~\ref{fig-Invariance} illustrates some running examples to show these properties. We discuss the examples using the {\em FTA by peak} version and set $k$ to $2$. The result is equally applicable to {\em FTA by thresholding}.

Following the notations in Section 3.1, we use $\mathbf X$, $\mathbf X'$ to denote two video sequences of the same action class. Two postures $l_1, l_2$ are obtained by projecting both video sequences into the corresponding subspaces, and we apply FTA by peak to encode the temporal order of these postures.  We use $S_{l_1}$, $S_{l_2}$ and $S'_{l_1}$, $S'_{l_2}$ to denote the confidence scores of posture $l_1$ and $l_2$ on $\mathbf X$, $\mathbf X'$, respectively. The occurrence of the peaks are at time $i_{l_1}, i_{l_2}$ for video $\mathbf X$, and $i'_{l_1}, i'_{l_2}$ for video $\mathbf X'$.

{\noindent\bf Temporal Translation Invariance} In Figure~\ref{fig-Invariance}(a), although the peaks of postures $l_1$ (red) and $l_2$ (blue) are at different locations for $\mathbf X$ and $\mathbf X'$ due to the temporal translation, the relationship $i_{l_1} < i_{l_2}$ and $i'_{l_1} < i'_{l_2}$ are consistent for these two video of the same class. The FTA hashing produces the code 1 for both videos.

{\noindent\bf Motion Scale Invariance} In Figure~\ref{fig-Invariance}(b), although $S_{l_1}$, $S_{l_2}$ (solid line) have a larger scale than $S'_{l_1}$, $S'_{l_2}$ (dash line), their peak order remains the same and the FTA hashing produces the same hash code for the videos of the same class.

{\noindent\bf Execution Rate Invariance} In Figure~\ref{fig-Invariance}(c), $S_{l_1}$ and $S_{l_2}$ (solid line) are squeezed due to the execution rate variation. For example, some people perform the action faster than the other people. However, the order of $i_{l_1} < i_{l_2}$ and $i'_{l_1} < i'_{l_2}$ are not affected by this variations and the FTA hashing produces the same code for $\mathbf X$ and $\mathbf X'$.


\section{Experiments}
In this section, we demonstrate the experiment results on several 3D action video datasets.

\subsection{Baseline Method}
As performing the hashing on the video level is a new problem, to our best knowledge, there is no existing methods in literature that can serve as the baseline. Thus we introduce a ``Bag-of-Words (BOW)" style method as the baseline for the comparison. The method also employs the random projection to each frame to produce a sequence of latent postures. However, it views each video as a bag and each posture as an item in the bag. In other words, it does not consider any temporal orders between the postures. Specifically, the BOW algorithm performs a thresholding test to decide whether an item of posture $l$ exists in the video, i.e., the hash code $i_l$ is given as
$$
  i_l=\begin{cases}
    0, & \text{if $\max\limits_{1\leq i\leq n} S_{l,i}<\theta$},\\
    1, & \text{otherwise}.
  \end{cases}
$$
where $\theta$ is a threshold to detect the existence of the posture $l$ according to its confidence score $S_{l,:}$ in the sequence. The above process is iterated to generate a sequence of binary hash bits. Since the BOW method does not leverage any temporal information, it can serve as a baseline method to validate the advantage of the temporal order-preserving FTA hashing algorithm.

\subsection{Feature Extraction}
Since the feature extraction is not the contribution of the current paper, we adopt the following four types of features commonly used in 3D human action recognition tasks.

\begin{enumerate}
\item \textbf{Pairwise-joint distance (PJD):} the normalized distance between a pair of joints \cite{wang2012mining, zhao2013online};
\item \textbf{Joint offset feature (JO):} the normalized joint offset from two consecutive frames \cite{orderlet14};
\item \textbf{Pairwise-angle feature (PA):}  the cosine of the angle between a pair of body segments \cite{ICMR2015}; and
\item \textbf{Histogram of Velocity Components (HVC):} the histogram of the 3D velocity of the point cloud in the neighborhood of the joints \cite{ICMR2015}.
\end{enumerate}

\subsection{Experiment Setting}
The parameter $\theta$ for the BOW, both versions of FTAs are chosen by the $5$-fold cross validation on the training set. Considering the proposed hashing algorithm is based on random projection and selection of postures, we repeat the experiment for $50$ runs and report the average accuracy as the results in all experiments. The Hamming distance is used as the distance metric for the KNN search to predict the label of an unknown test sequence.

\subsection{Experiment Results}
We conduct performance evaluations on public three mostly used 3D action video datasets. It is worth noting that the proposed FTA hashing algorithm is especially amenable to large-scale tasks, however, to our best knowledge, there is no extremely large 3D action datasets publicly available in literature.  But the results demonstrated on these datasets should suffice to show the competitive performance of the proposed algorithm.

\subsubsection*{UTKinect-Action Dataset}
The UTKinect-Action dataset \cite{UTKinect} consists of $10$ action types performed by 10 subjects. All subjects perform each action twice. Since subjects are free to move in the environment, the dataset is very challenging due to the huge viewpoint variation and intra-class variance. We follow the same cross-subject test setting from \cite{2014Liegroup}.

Experimental results are summarized in Table~\ref{Tab:UTcmp} in which we compare the recognition accuracy of the BOW, the FTA by peak and the FTA by thresholding with four types of features. The $(k+1)$-ary code length $p$ is set to $1,000$, $k$ is set to $2$.  The FTA by peak produces the accuracy of $90.20\%$ and $86.57\%$ on the PA feature and the PJD feature, respectively. This is a very impressive performance considering that we only use the hashing and the approximate nearest neighbor search by the Hamming distance. As shown, both the FTA by peak and the FTA by thresholding outperform the baseline BOW method by more than $10\%$, demonstrating the contribution of the modeling temporal order to performance of the FTA hashing. We also note that FTA by peak has a higher performance than the FTA by thresholding. This is probably because the peak is a more robust estimate of occurrence time of a posture than the onset time passing a confidence threshold. 

It is also worth noting that the performances vary across different features. The PA feature achieves an accuracy of $90.2\%$ while the HVC feature has only $69.6$ in accuracy. This is normal because the discriminative capabilities of different features are different. We report the accuracy of multiple features to show the performance of FTA can consistently outperform the BOW. The recognition accuracy can be further boosted by fusing multiple features  (e.g. concatenation) but this is out of the scope of this paper.

\begin{table}
\begin{center}
\begin{tabular}{| >{\centering\arraybackslash}m{0.35in} | >{\centering\arraybackslash}m{0.78in} |>{\centering\arraybackslash}m{0.78in} |>{\centering\arraybackslash}m{0.78in} |}
\hline
 Features & BOW &FTA by Thresholding  & FTA by Peak \\
 \hline
PJD & $76.97 \pm 1.63$ & $84.44 \pm 2.29$& $\mathbf{86.57 \pm 2.52}$ \\
\hline
JO & $72.02 \pm 2.65$  & $71.01 \pm 2.23$& $73.43 \pm 2.07$ \\
\hline
PA & $78.28\pm 2.98$ &$85.76 \pm 1.81$ & $\mathbf{90.20 \pm 1.43}$ \\
\hline
HVC & $59.69 \pm 1.46$ & $66.06 \pm 1.18$ & $69.60 \pm 1.93$ \\
\hline
\end{tabular}
\end{center}
\caption{Performance comparison between different hashing methods on the UTKinect-Action dataset ($k=2$, $p=1000$, accuracy in terms of \%).}
\label{Tab:UTcmp}
\end{table}

\begin{figure}
\begin{center}
   \includegraphics[width=0.8\linewidth]{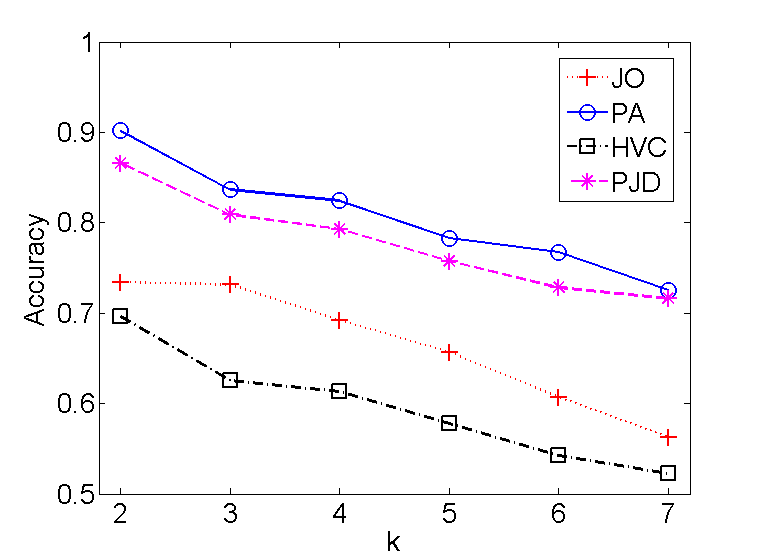}
\end{center}
   \caption{Relationship between $k$ and the accuracy on different features on the UTKinect-Action dataset ($(k+1)$-ary code length is set to 1000).}
\label{fig-utK}
\end{figure}

Next, we further study the performance with respect to different $k$ (i.e., the $(k+1)$-ary code used by FTA) as well as the threshold $\theta$. We only report the result on the FTA by peak since it is consistently better than FTA by thresholding. Figure~\ref{fig-utK} shows the accuracy versus $k$ when the code length is set to $1000$. The accuracy drops when $k$ increases, where $k=2$ gives the best performance. This is because a larger $k$ may produce redundant codes when comparing a large number of postures -- postures may be dominated by a few more salient postures resulting in the loss of the discriminative information. 

\begin{figure}
\begin{center}
   \includegraphics[width=0.8\linewidth]{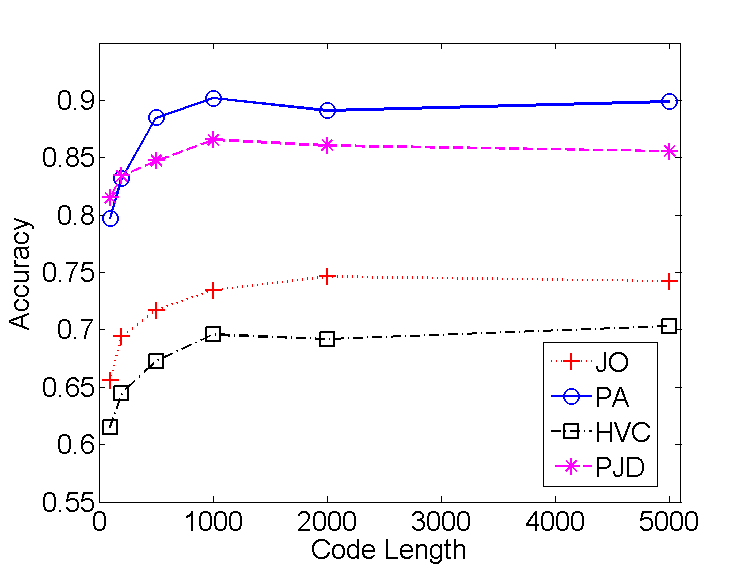}
\end{center}
   \caption{Relationship between the $(k+1)$-ary code length and the accuracy on different features on the UTKinect-Action dataset ($k$ is set to 2).}

\label{fig-utL}
\end{figure}

Figure~\ref{fig-utL} shows the effect of code length on the performance when $k=2$. The accuracy increases while the code length grows.  The accuracy remains stable after the code length reaches $1,000$.  Note that the code length is with respect to the $(k+1)$-ary code. In other words, we use $\lceil\log (k+1)\rceil$ binary bits to encode a $(k+1)$-nary code.

As shown, the entire video requires only $100$ $(k+1)$-ary hash codes (i.e., $200$ bits of codes when $k=2$) to achieve an accuracy higher than $80\%$ with PJD and PA features.  {\bf Suppose if a video has $\mathbf {50}$ frames, it means we only need $\mathbf 4$ bits per frame}, which is extremely efficient considering state-of-the-art supervised hashing methods for the image retrieval normally needs $32 \sim 64$ bits for a single image to achieve a comparable performance \cite{Kai_Arxiv}. This shows significant efficiency the proposed FTA can achieve.

\begin{figure}
\begin{center}
   \includegraphics[width=0.8\linewidth]{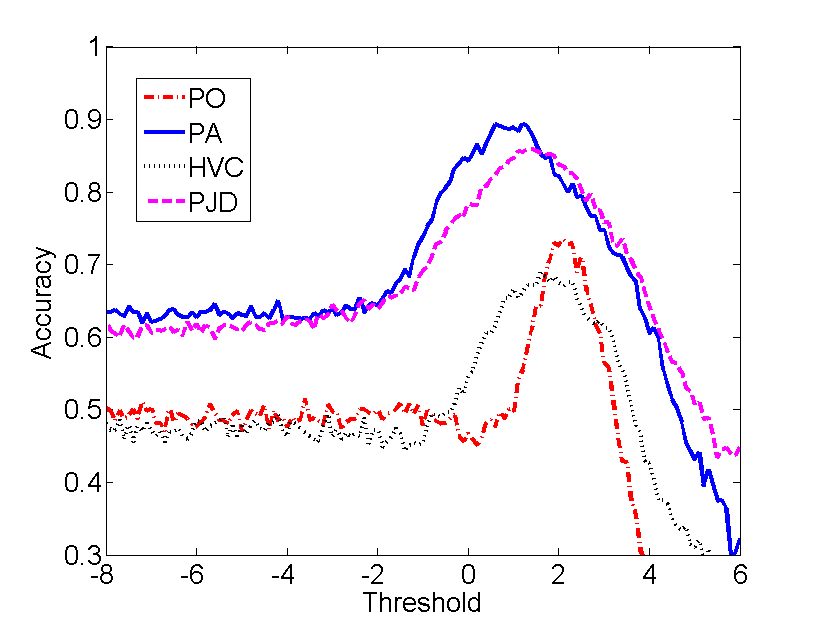}
\end{center}
   \caption{Relationship between the threshold and the accuracy on different features on the UTKinect-Action dataset ($k$ is set to 2 and $(k+1)$-ary code length is set to 1000).}
\label{fig-utThres}
\end{figure}

Finally, we also show the effect of the threshold on the performance of the FTA by peak. In this experiment, $k$ is fixed to 2, the $(k+1)$-ary code length is set to $1,000$. In Figure~\ref{fig-utThres}, consistent performance has been shown on accuracy versus threshold across all four features. When the threshold is set to a small value, many false postures can pass the threshold test. On the contrary, when the threshold is set to a larger value, true postures can fail the test resulting a hash code full of "0" bits. Both cases would compromise the accuracy. Thus the threshold should be set in a reasonable range for the satisfactory level of accuracy by avoiding false positive or false negative detection of latent postures.

\subsubsection*{MSR Action3D Dataset}
The MSR Action3D dataset \cite{li2010action} covers $20$ sports action types and 10 subjects. All subjects perform each action two or three times. The dataset is very challenging due to the high intra-class variations. We follow the same experiment settings in \cite{wang2012mining}.

Similar to the experiment in the previous UTKinect-Action dataset, we compare the performance among the baseline BOW, FTA by threshold and the FTA by peak. Results are reported in Table~\ref{Tab:Action3Dcmp}. Again, FTA by peak has achieved the best accuracy than the other two methods across all features.  Different features have produced different level of accuracies. The JO feature achieves an accuracy of $77.81\%$ while the PJD feature has only $50.23\%$ in accuracy. 


\begin{table}
\begin{center}
\begin{tabular}{| >{\centering\arraybackslash}m{0.4in} | >{\centering\arraybackslash}m{0.78in} |>{\centering\arraybackslash}m{0.78in} |>{\centering\arraybackslash}m{0.78in} |}
\hline
 Features & BOW &FTA by Thresholding  & FTA by Peak \\
 \hline
PJD & $44.79 \pm 2.15$ & $44.02 \pm 1.60$& $50.23 \pm 1.09$ \\
\hline
JO & $50.69 \pm 1.92$  & $58.97 \pm 1.38$& $\mathbf{77.81 \pm 2.25}$ \\
\hline
PA & $49.58\pm 1.29$ &$51.65 \pm 1.05$ & $55.94 \pm 2.12$ \\
\hline
HVC & $49.27 \pm 2.86$ & $50.54 \pm 1.34$ & $57.32 \pm 2.27$ \\
\hline
\end{tabular}
\end{center}
\caption{Performance comparison between different hashing methods on the MSR Action3D dataset ($k=2$, $p=1000$, accuracy in terms of \%).}
\label{Tab:Action3Dcmp}
\end{table}

\begin{figure}
\begin{center}
   \includegraphics[width=0.8\linewidth]{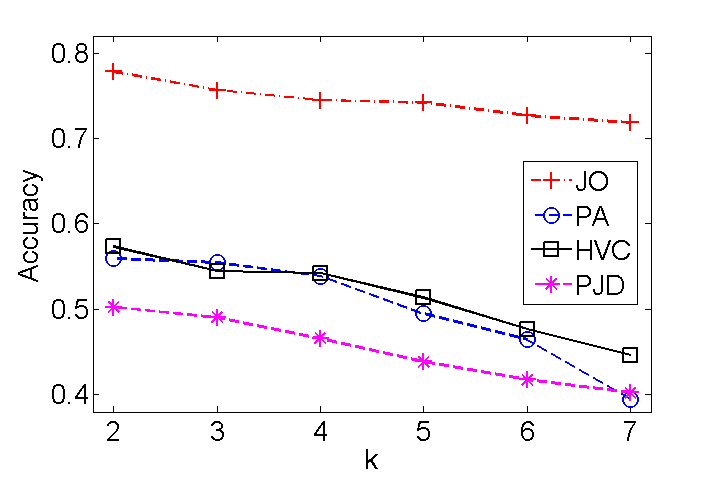}
\end{center}
   \caption{Relationship between $k$ and the accuracy on different features on the MSR Action3D dataset. ($(k+1)$-ary code length is set to 1000)}
\label{fig-3DK}
\end{figure}

\begin{figure}
\begin{center}
   \includegraphics[width=0.8\linewidth]{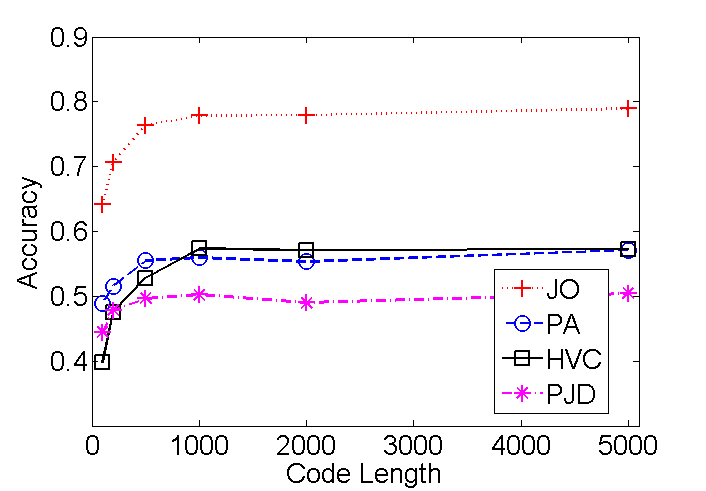}
\end{center}
   \caption{Relationship between the $(k+1)$-ary code length and the accuracy on the MSR Action3D dataset.}
\label{fig-3DL}
\end{figure}

We also evaluate the impact of $k$ and code length on the accuracy of the FTA by peak algorithm, and show the results in Figure~\ref{fig-3DK} and Figure~\ref{fig-3DL}. As shown, the recognition accuracy drops as $k$ increases, while a longer code length usually produces higher accuracy. This results are consistent with the results on the MSR Action3D dataset.

\subsubsection*{MSRActionPairs Dataset}
The MSRActionPairs dataset \cite{hon4d} consists of $12$ action types performed by $10$ subjects. Each subject performs every action three times. This dataset contains 6 pairs of similar actions which has exactly the same poses but different temporal orders. For example, "Pick up" and "Put down", "Push a chair" and "Pull a chair". This dataset is very suitable to evaluate the temporal order-preserving capability of the proposed FTA hashing algorithm. We follow the same test setting of \cite{hon4d}.

Table~\ref{Tab:ActionPairscmp} compares the results among the FTA by peak, FTA by thresholding and the baseline BOW algorithm. The recognition accuracy of the FTA by peak significantly outperforms the BOW algorithm by $15 \sim 20\%$ on most of the features. Since the MSRActionPairs dataset is very sensitive to the temporal order of the action sequences, the FTA hashing can effectively distinguish between different actions with similar postures but in different temporal orders. On the contrary, BOW does not encode the temporal structure, and is incapable of handling this challenging setting on this dataset. 

\begin{figure}
\begin{center}
   \includegraphics[width=0.8\linewidth]{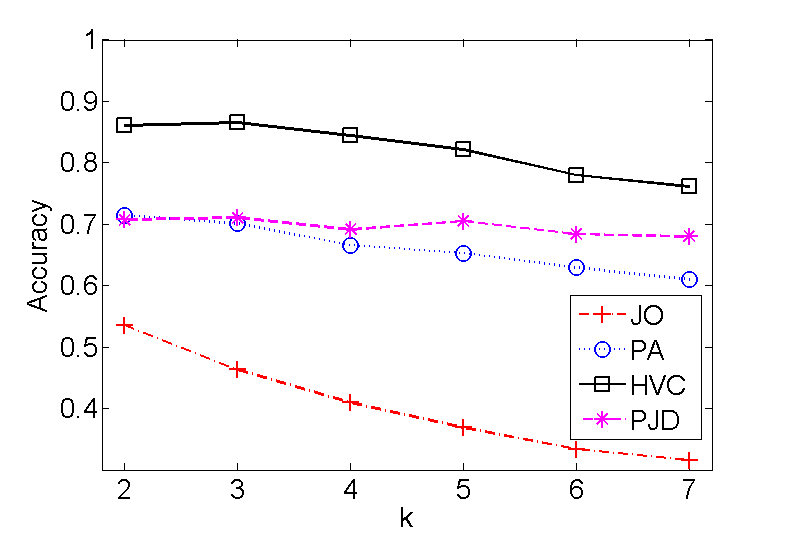}
\end{center}
   \caption{Relationship between $k$ and the accuracy on the MSRActionPairs dataset.}
\label{fig-pairsK}
\end{figure}

\begin{figure}
\begin{center}
   \includegraphics[width=0.8\linewidth]{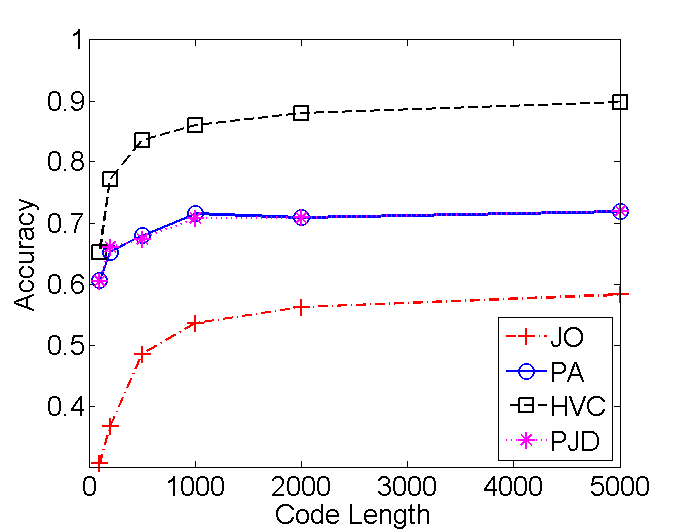}
\end{center}
   \caption{Relationship between the $(k+1)$-ary code length and the accuracy on the MSRActionPairs dataset.}
\label{fig-pairsL}
\end{figure}

In addition, we perform the same set of experiments on the impact of $k$ and the code length on the performance in Figure~\ref{fig-pairsK} and Figure~\ref{fig-pairsL}. Similar results are observed as for the other two datasets.

\begin{table}
\begin{center}
\begin{tabular}{| >{\centering\arraybackslash}m{0.4in} | >{\centering\arraybackslash}m{0.78in} |>{\centering\arraybackslash}m{0.78in} |>{\centering\arraybackslash}m{0.78in} |}
\hline
 Features & BOW &FTA by Thresholding  & FTA by Peak \\
 \hline
PJD & $50.74 \pm 1.95$ & $59.71 \pm 2.28$& $70.74 \pm 1.72$ \\
\hline
JO & $50.40 \pm 3.47$  & $57.25 \pm 2.73$& $53.6 \pm 1.72$ \\
\hline
PA & $57.37\pm 1.32$ &$61.48 \pm 1.18$ & $71.48 \pm 2.26$ \\
\hline
HVC & $64.11 \pm 3.48$ & $75.20 \pm 2.20$ & $\mathbf{86.05 \pm 2.61}$ \\
\hline
\end{tabular}
\end{center}
\caption{Performance comparison between different hashing methods on the MSRActionPairs dataset (Accuracy in terms of \%).}
\label{Tab:ActionPairscmp}
\end{table}

\section{Conclusions}

In this paper, we revisit the human action recognition problem from a hashing perspective and propose a novel First-Take-All hashing algorithm to interpret the temporal patterns of the entire video. The FTA hashing preserves the temporal order of the action sequence and achieves invariance to the temporal translation, motion scale as well the execution rate. Experiment results on three public 3D human action datasets have demonstrated the performance and the efficiency of the proposed FTA hashing.

The current work is based on random projection. We would like to further enhance the performance of the FTA hashing and shrink the code length by leveraging the learning-based method in our future work.

{\small
\bibliographystyle{ieee}
\bibliography{iccv2015}
}

\end{document}